\documentclass{article}

\usepackage[mlss]{cosparws_abs} 

\usepackage[utf8]{inputenc} 
\usepackage[T1]{fontenc}    
\usepackage{hyperref}       
\usepackage{url}            
\usepackage{booktabs}       
\usepackage{amsfonts}       
\usepackage{nicefrac}       
\usepackage{microtype}      
\usepackage{graphicx}       
\usepackage{algorithm}
\usepackage{algorithmic}

\title{Scalable Data Balancing \\ 
for Unlabeled Satellite Imagery}

\author{
  Deep Patel\thanks{Equal Contribution; Work done as researchers at Space ML \cite{Ganju2020FDL}} \\
  \texttt{dmpatel@andrew.cmu.edu} \\
  \And
  Erin Gao\footnotemark[1] \\ 
  \texttt{eringao@andrew.cmu.edu} \\
  \And
  Anirudh Koul\footnotemark[1] \\
  \texttt{anirudhkoul@gmail.com} \\
  \AND
  Siddha Ganju \\
  
  \And 
  Meher Anand Kasam \\
}

\begin{document}

\maketitle

\section{Introduction}
Data imbalance is a ubiquitous problem in machine learning. In large scale collected and annotated datasets, data imbalance is either mitigated manually (by undersampling frequent classes and oversampling rare classes), or planned for with imputation and augmentation techniques. In both cases balancing data requires labels. In other words, only annotated data can be balanced. Collecting fully annotated datasets is challenging, especially for large scale satellite systems such as the unlabeled NASA’s 35 PB Earth Imagery dataset \cite{seeley2020kdf}. 
\\ 
Although the NASA Earth Imagery dataset is unlabeled, there are implicit properties of the data source that we can rely on to hypothesize about its imbalance - such as distribution of land and water in the case of Earth’s imagery. Since over 70\% of Earth’s surface is covered by water bodies, large scale satellite data is expected to have a much higher percentage of water imagery. On the other hand, phenomena like hurricanes would have rare occurrences. A naive ML model can classify everything as “ocean” and still achieve a 70\% accuracy. 
\\ 
We present a new iterative method to balance unlabeled data - our method utilizes image embeddings as a proxy for image labels that can be used to balance data, and ultimately when trained increases overall accuracy.

\section{Diversely Selecting Training Samples}

Starting with unbalanced and unlabeled data, we use self-supervised learning methods (including autoencoder and SimCLR) to train a model, then use the trained model (an embedder) to generate embeddings for the entire dataset. Clusters are identified based on these embeddings and an equal number of points per cluster are chosen to generate a new balanced training dataset.
\newline
Again, a self-supervised model is trained with this dataset which is further used to generate a more balanced training dataset. Each iteration improves the representative power of the embeddings, thereby improving the definition of the next cluster, balancing data further, and interestingly, leads to much higher accuracy on the same dataset size compared to a random dataset. Our experiments on a sample unbalanced dataset which when balanced by the algorithm gives 90\% accuracy, a considerable 50\% improvement over a baseline with 40\% accuracy and 2.5x the number of training images. 
\newline
The method presented is generalizable and can be extended beyond image modalities to text and video data where data imbalance is similarly ubiquitous. Since our annotation technique is largely automatic, our method is scalable and generalizable and especially useful for unexplored fields with abundant unlabeled data.
\newpage

\begin{algorithm}
\caption{Diversely Select N Embeddings from \it embedder, \it data, \it n}
\begin{algorithmic}
\STATE $features \leftarrow embedder(data)$
\STATE $result \leftarrow [random.choice(data)]$
\STATE $distances \leftarrow [] * len(data)$
\WHILE{$n \neq 0$}
\STATE $i \leftarrow 0$
\WHILE{$i < len(features)$}
\STATE $dist \leftarrow dist(features[i], result[-1])$
\IF{$distances[i] > dist$}
\STATE $distances[i] = dist$
\ENDIF
\ENDWHILE
\STATE $idx \leftarrow distances.index(max(distances))$
\STATE $result.append(data[idx])$
\STATE $del features[idx]$
\STATE $n \leftarrow n - 1$
\ENDWHILE
\STATE $return \ result$
\end{algorithmic}
\end{algorithm}

\section{Experiments}

To understand the benefits of selecting embeddings with this iterative method, we analyze the benefits with experiments on two different datasets. With a Baised Circles Dataset, we analyze how selecting 500 training points with this method improves model accuracy over a baseline of selecting points randomly. With a biased sampling of the UC Merced satellite imagery dataset, we view how diversely selecting training samples improves model accuracy, especially as this is process is done iteratively over multiple model iterations.

\subsection{Circles Dataset}
500 points were selected, via two different methods, from the Biased Circles dataset (Figure 1) above. The dataset is composed of 4500 points, which consists of 10 rings. With each circle numbered from 1 to 10 from outer circle to inner circle, each odd numbered circle containing eight times as many elements as its adjacent inner circle. The colors of the rings correlate with which class the points belong to where each circle is its own class of points

\begin{figure}[h]
    \centering
    \includegraphics[width=0.50\textwidth]{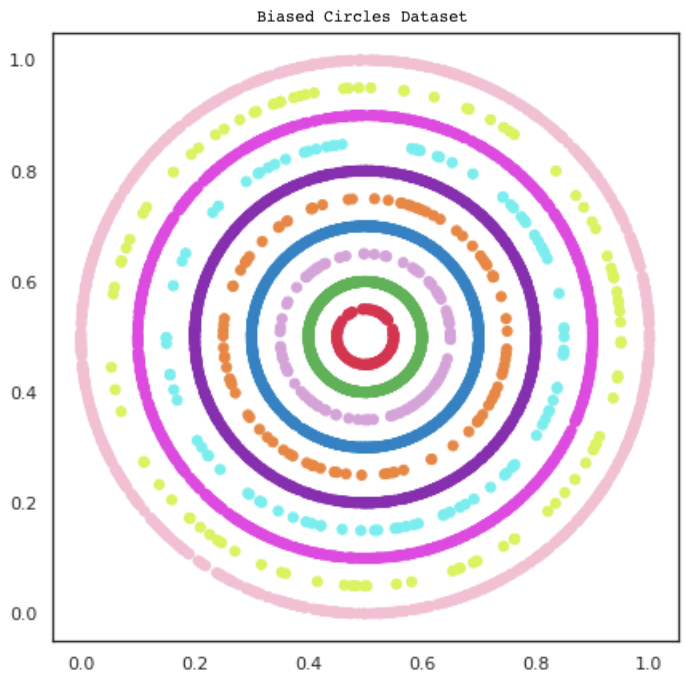} 
    \caption{Biased Circles Dataset}
\end{figure}

\newpage 

Figure 2 is an image of the data points selected randomly whereas Figure 3 consists of data points chosen using diverse embedding selection. The latter set of data points are far more structured than the randomly selected points as they do not retain nearly as much of the imbalance in the original dataset.

\begin{figure}[h]
    \centering
    \begin{minipage}{0.50\textwidth}
        \centering
        \includegraphics[width=0.90\textwidth]{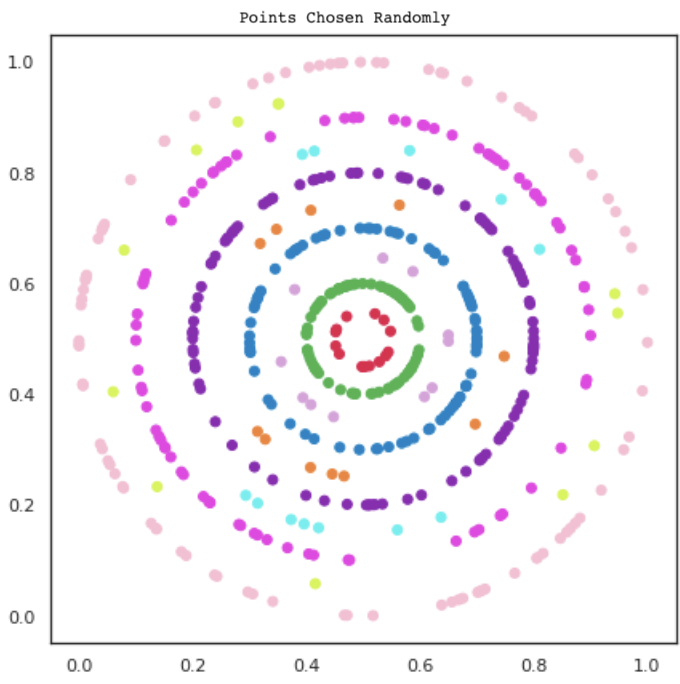} 
        \caption{Randomly chosen embeddings}
    \end{minipage}\hfill
    \begin{minipage}{0.50\textwidth}
        \centering
        \includegraphics[width=0.90\textwidth]{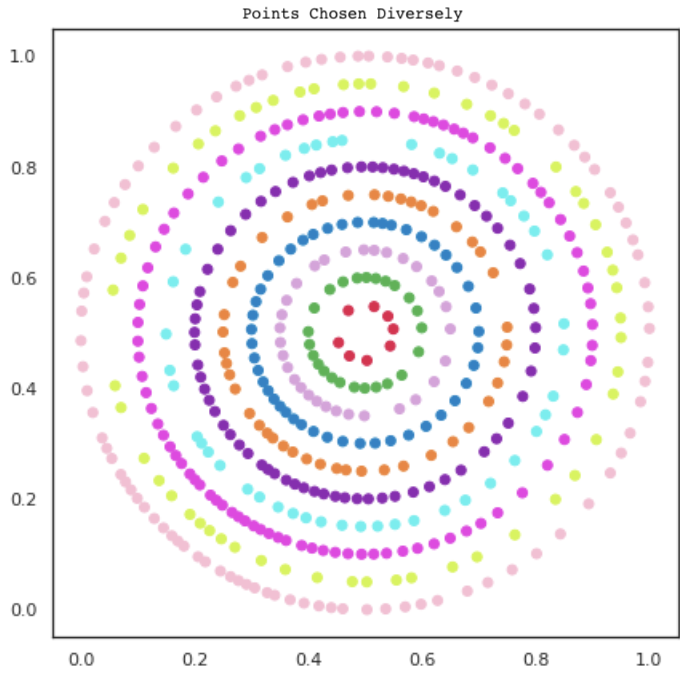} 
        \caption{Diversely chosen embeddings}
    \end{minipage}
\end{figure}

After classifying on a randomly selected test set, it can be seen in Figure 4 that diversely chosen points provide a little over a 40\% increase in model accuracy from the baseline model that was trained on randomly chosen training points which is a very significant improvement.

\begin{figure}[h]
    \begin{center}
    \begin{tabular}{|| c | c | c ||}
    \hline
    Selection Method & No. Training Points & Avg Accuracy \\
    \hline\hline
    Random & $500$ & $0.484 \pm 0.003$ \\
    \hline
    Diverse & $500$ & $0.906 \pm 0.011$ \\
    \hline
    \end{tabular}
    \caption{Training Model on 500 Points}
    \end{center}
\end{figure}

\begin{figure}[h]
    \centering
    \begin{minipage}{0.45\textwidth}
        \centering
        \includegraphics[width=0.80\textwidth]{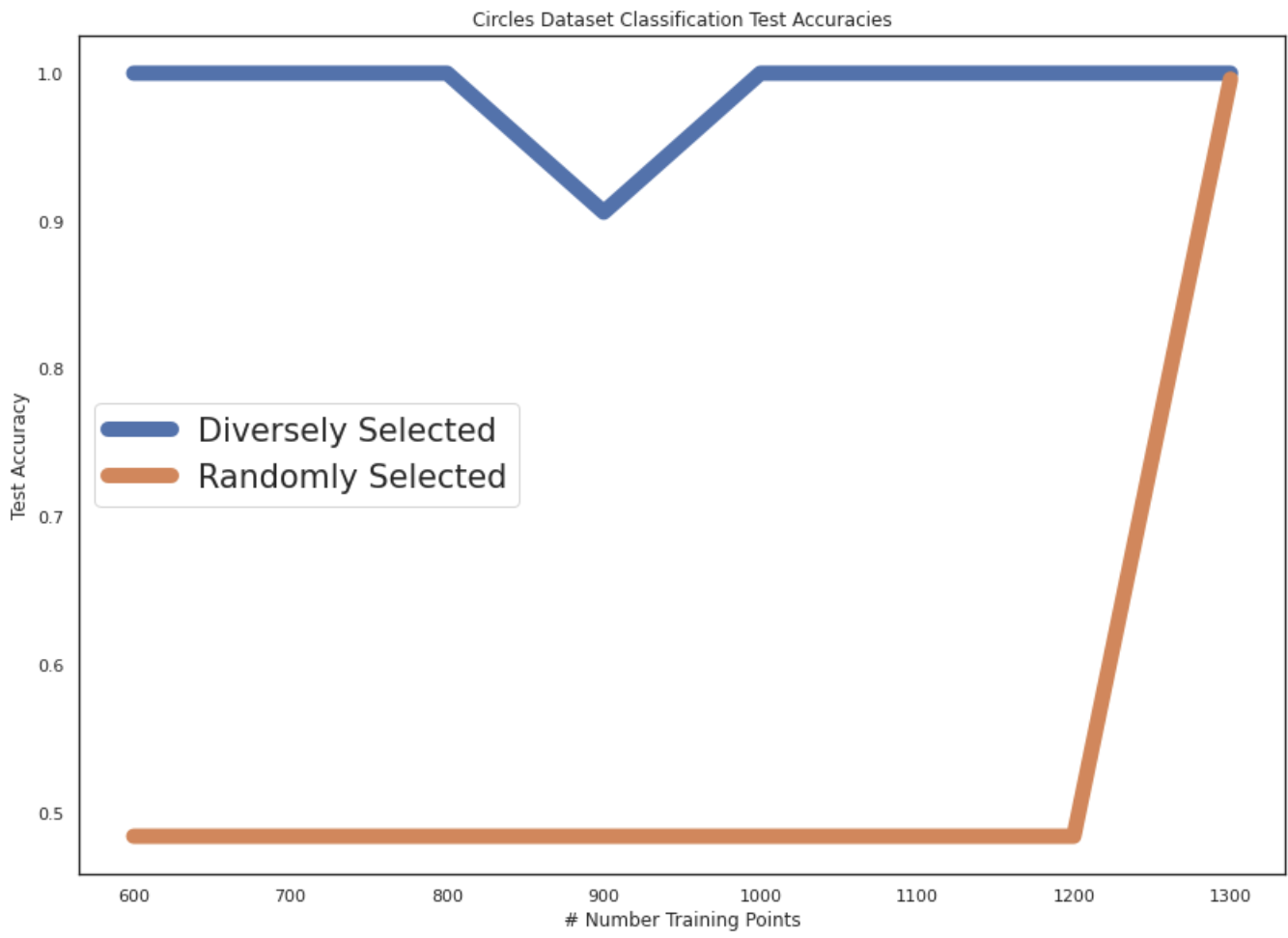}
        \caption{Classification Accuracies for Larger Training Sets}
    \end{minipage}\hfill
    \begin{minipage}{0.45\textwidth}
        \centering
        \includegraphics[width=0.80\textwidth]{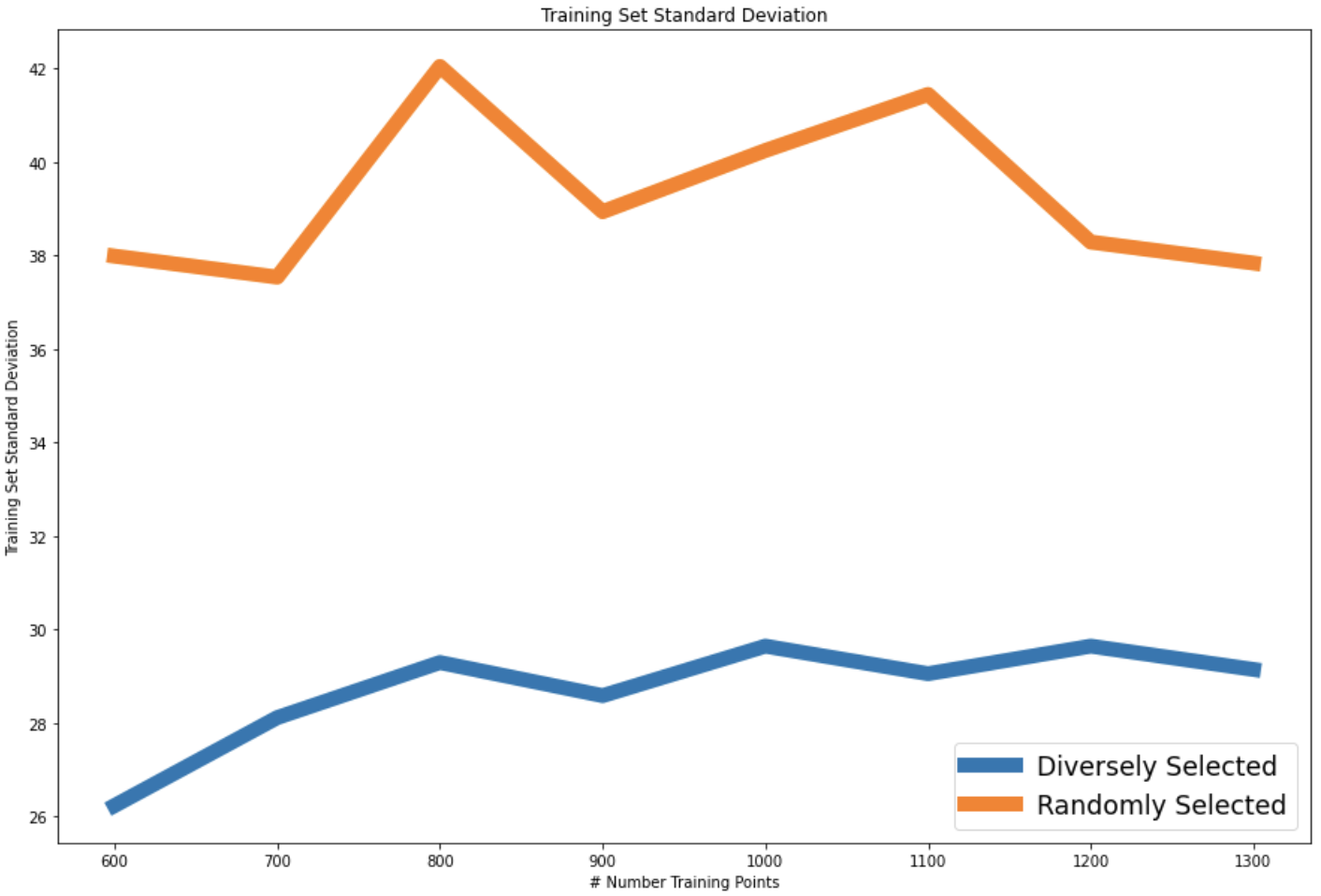}
        \caption{Class-size Standard Deviations from Selected Training Sets}
    \end{minipage}\hfill
\end{figure}

Furthermore, in Figure 5, we see that these results are consistent when the size of the selected training points increase. The difference is mitigated when the difference between the random and diversely selected training sets is small at larger trainings sets. However, it takes about 250\% more randomly selected training data to reach a trained model accuracy as good as one trained on a diversely selected training set. This is expected as more points selected makes the diversely selected points more similar to the randomly selected points. Regardless, on the circles dataset, it can be seen that when training on smaller datasets, diversly selected training sets can greatly improve model performance. In Figure 6, it can be seen that selecting features using this strategy allows for a training dataset that is more balanced compared to the the randomly selected training set. This is noticable because the standard deviation between the classes is far less for diversly-selected training sets. This means the number of features in each training set are far closer than the randomly selected dataset. This result shows that diversely selecting features allows for much better representation from each class in the overall training set.

\subsection{UC Merced Dataset}
Selecting training points diversely is primarily is intended to work best with unbalanced data; however, the UC Merced satellite imagery dataset is a balanced dataset. As a result, we subset this dataset into one that produces an unbalance across the UC Merced dataset classes. The dataset that we use in this experiment can be seen in Figure 7.

\begin{figure}[h]
    \centering
    \begin{minipage}{0.80\textwidth}
        \centering
        \includegraphics[width=0.90\textwidth]{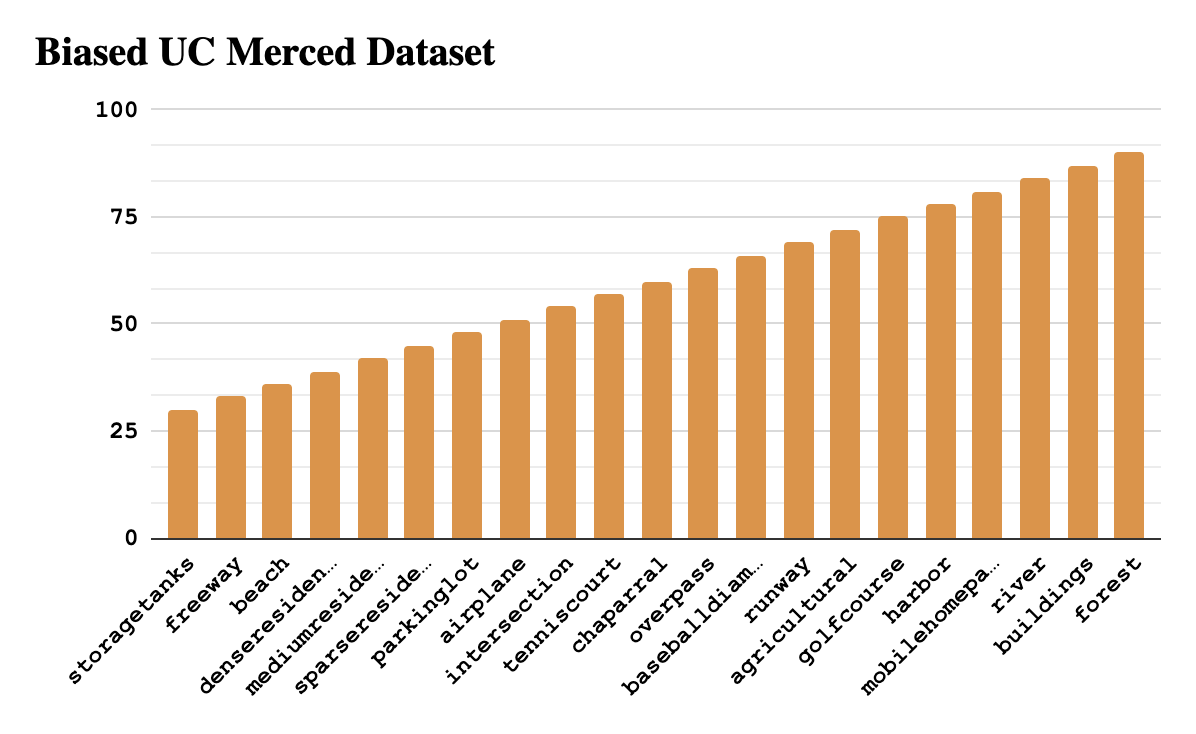}
        \caption{Unbalanced Subset of UC Merced Dataset}
    \end{minipage}\hfill
\end{figure}

For this experiment, we first train an autoencoder to generate a list of diverse or random features on the unbalanced dataset. Then, we iteratively train a model on the most these features, use the model to make an encoder, obtain diverse or random features, and repeat. At each model iteration, we expect the models to improve, produce a better encoding representation of the dataset, find a better balanced training set, and train an even better model. The results of the model performance and the class representation of the training set (represented by the standard deviation) over each model iteration can be found in Figure 8 and Figure 9, respectively.

\begin{figure}[h]
    \centering
    \begin{minipage}{0.45\textwidth}
        \centering
        \includegraphics[width=0.90\textwidth]{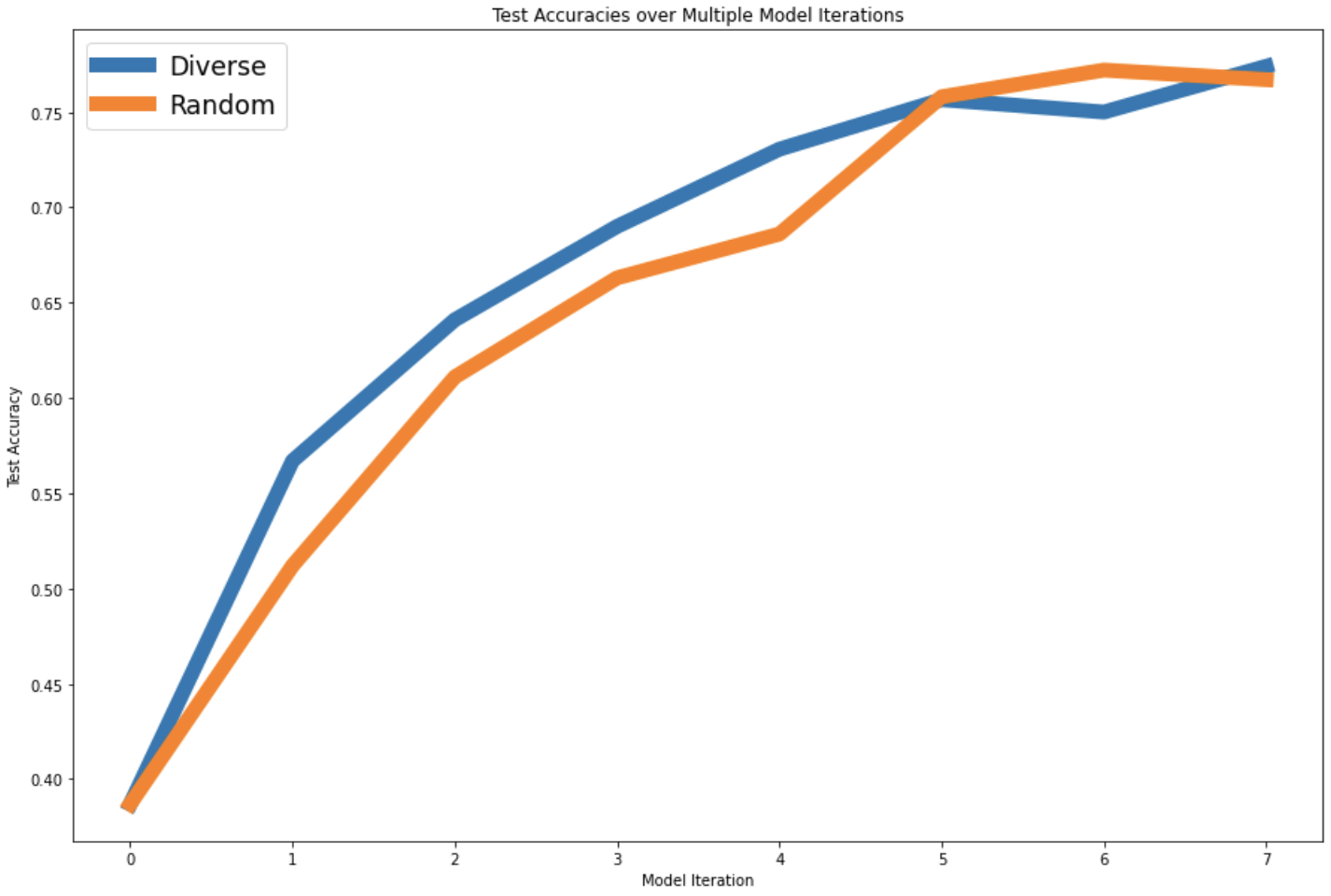}
        \caption{Training Accuracies from Iteratively Training Models}
    \end{minipage}\hfill
    \begin{minipage}{0.45\textwidth}
        \centering
        \includegraphics[width=0.90\textwidth]{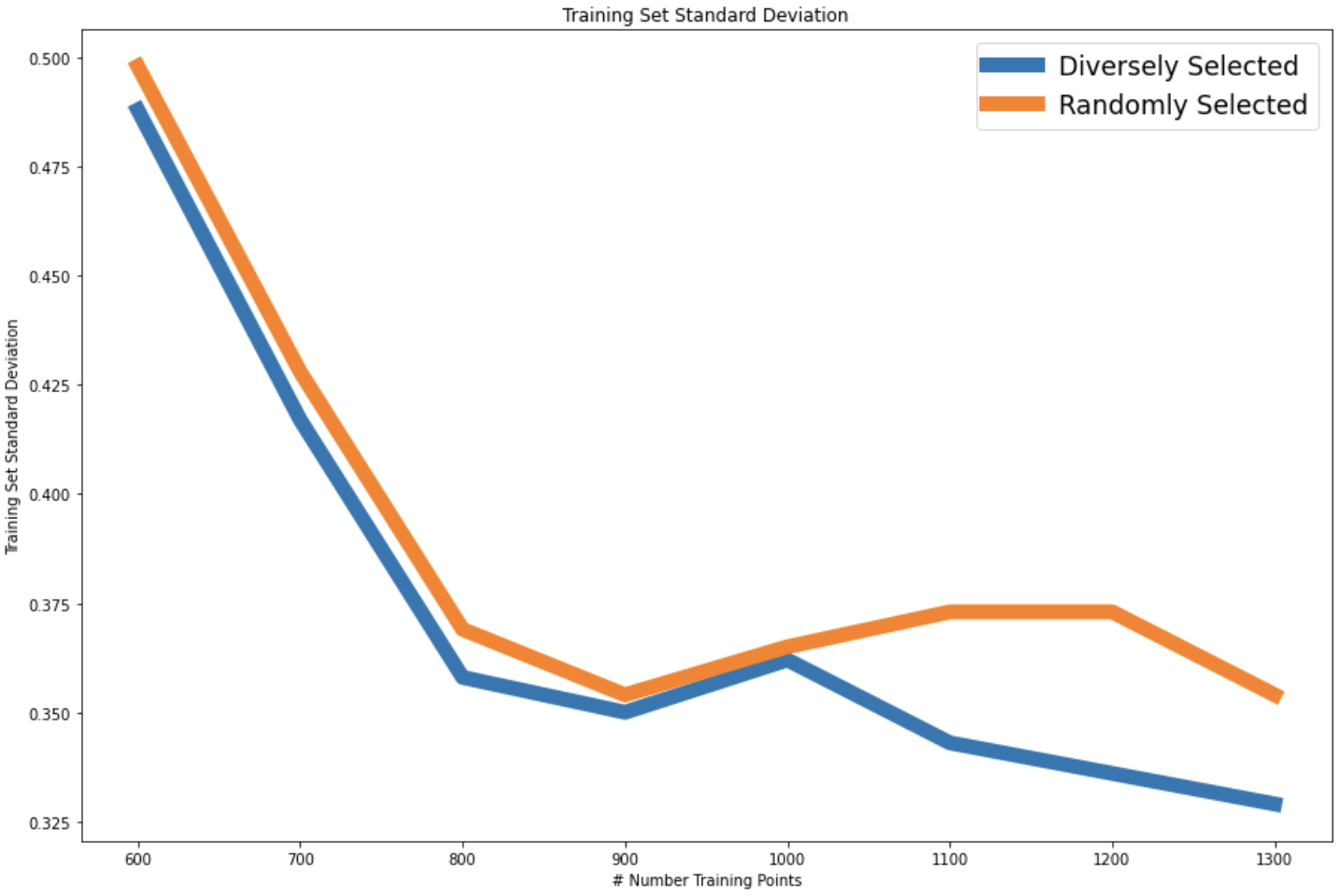}
        \caption{Class-size Standard Deviations from Selected Training Sets}
    \end{minipage}\hfill
\end{figure}

The results show that there is a small improvement in the model test accuracies throughout most of the iterations; however, this difference appears to be mitigated at later iterations. The results of this experiment show that diversely selecting features still allows for some improvement over models trained on randomly selected features. Similarly, the representation of classes between training sets is similar, but the diversly selected features have slightly better balance. It is important to notice that these improvements here are not as drastic as the experiments with the circles dataset. As a result, it is clear that the improvements from diversely selecting a training set can be dependent on the structure and features in the dataset.

\section{Conclusions}
As a result, through the experiments with the circles and UC Merced dataset, we see that the iterative method to create balanced data is viable and can provide great improvements to class representation in a dataset. As a summary, these are the high-level takeaways that can be obtained from these experiments:
\begin{itemize}
    \item The iterative method to diversely balance data is effective at balancing unbalaced data
    \item Using datasets that are produced from diversly selected features tends to result in better models due to the better balanced data
    \item The benefits of diversely selecting features is likely to be more dependent on the dataset and more effective on datasets with higher unbalances.
\end{itemize}
The results provide justification for a variety of many other experiments on the process of balancing unlabeled and unbalanced data.

The effectiveness of this iterative selection method has appeared to be contingent on the structure of the dataset as seen by the difference in the results of the experiments of circles and unbalanced UC Merced dataset. This may require furthur experimentation as it would be important to determine what datasets would be effective for using this method. For example, running similar experiments on varying datasets depending on feature variation, dataset size, and overall structure. 

Furthermore, varying the shape of the dataset across classes would provide insight into what kinds of datasets would benefit from this iterative method for feature selection. Other experiments include varying the structure and methods for training and creating the initial self-supervised model to obtain different representations of the feature data. 

Finally, this method was originally intended to work on larger datasets than those shown in the paper. As a result, researching a concurrent implementation of this algorithm to see how well this method can perform and balance extremely large datasets would shed light on the effectiveness in the practical use of this algorithm.

\bibliography{references}

\begin{thebibliography}{2}
\providecommand{\natexlab}[1]{#1}
\providecommand{\url}[1]{\texttt{#1}}
\expandafter\ifx\csname urlstyle\endcsname\relax
  \providecommand{\doi}[1]{doi: #1}\else
  \providecommand{\doi}{doi: \begingroup \urlstyle{rm}\Url}\fi

\bibitem[Ganju et~al.(2020)Ganju, Koul, Lavin, Veitch-Michaelis, Kasam, and
  Parr]{Ganju2020FDL}
Siddha Ganju, Anirudh Koul, Alexander Lavin, Josh Veitch-Michaelis, Meher
  Kasam, and James Parr.
\newblock {Learnings from Frontier Development Lab and SpaceML - AI
  Accelerators for NASA and ESA}, 2020.

\bibitem[Seeley et~al.(2020)Seeley, Civilini, Srishankar, Praveen, Koul, Berea,
  and El-Askary]{seeley2020kdf}
M~Seeley, F~Civilini, N~Srishankar, Satyarth Praveen, Anirudh Koul, Anamarea
  Berea, and Hosam El-Askary.
\newblock {Knowledge Discovery Framework}, 2020.

\end{thebibliography}
\bibliographystyle{plainnat}

\end{document}